\documentclass[10pt,twocolumn,letterpaper]{article}

\usepackage{iccv}              

\usepackage{graphicx}
\usepackage{multirow}
\usepackage{wrapfig}

\definecolor{iccvblue}{rgb}{0.21,0.49,0.74}
\usepackage[pagebackref,breaklinks,colorlinks,allcolors=iccvblue]{hyperref}
\usepackage[accsupp]{axessibility}

\def\paperID{2266} 
\def\confName{ICCV}
\def\confYear{2025}
\def\nickName{MuGS}

\title{\nickName: Multi-Baseline Generalizable Gaussian Splatting Reconstruction}

\author{
    Yaopeng Lou \quad
    Liao Shen \quad
    Tianqi Liu \quad
    Jiaqi Li \\
    Zihao Huang \quad
    Huiqiang Sun \quad
    Zhiguo Cao$^\dagger$
    \vspace{1pt} \\
    School of AIA, Huazhong University of Science and Technology\\
}

\begin{document}
\maketitle
\begin{abstract}
We present Multi-Baseline Gaussian Splatting (\nickName), a generalized feed-forward approach for novel view synthesis that effectively handles diverse baseline settings, including sparse input views with both small and large baselines.
Specifically, we integrate features from Multi-View Stereo (MVS) and Monocular Depth Estimation (MDE) to enhance feature representations for generalizable reconstruction. Next, We propose a projection-and-sampling mechanism for deep depth fusion, which constructs a fine probability volume to guide the regression of the feature map. Furthermore, We introduce a reference-view loss to improve geometry and optimization efficiency. We leverage $3$D Gaussian representations to accelerate training and inference time while enhancing rendering quality.
\nickName\ achieves state-of-the-art performance across multiple baseline settings and diverse scenarios ranging from simple objects (DTU) to complex indoor and outdoor scenes (RealEstate10K). We also demonstrate promising zero-shot performance on the LLFF and Mip-NeRF 360 datasets. Code is available at \href{https://github.com/EuclidLou/MuGS}{https://github.com/EuclidLou/MuGS}.
\end{abstract}
\renewcommand\thefootnote{} %
\footnotemark
\footnotetext{$^\dagger$Corresponding author.}
\renewcommand\thefootnote{\arabic{footnote}} %
\section{Introduction}
\label{sec:intro}

Novel view synthesis (NVS) represents a fundamental and practical challenge in computer vision and graphics. Neural Radiance Fields (NeRF) \cite{mildenhall_nerf_2020}, which encode scenes as implicit radiance fields, have demonstrated remarkable success. However, NeRF is computationally expensive, as it requires querying dense points for rendering. Recently, 3D Gaussian Splatting (3D-GS) \cite{kerbl_3d_2023} has emerged as an efficient alternative, leveraging anisotropic 3D Gaussians to represent scenes explicitly. This approach facilitates real-time, high-quality rendering through a differentiable tile-based rasterizer. Despite these advances, 3D-GS requires per-scene optimization, which remains time-consuming, limiting its practical applicability.

\begin{figure}[thbp]
	\centering
    \begin{minipage}[c]{\linewidth}
	\begin{minipage}[c]{0.475\linewidth}
		\centering
		\includegraphics[width=\textwidth]{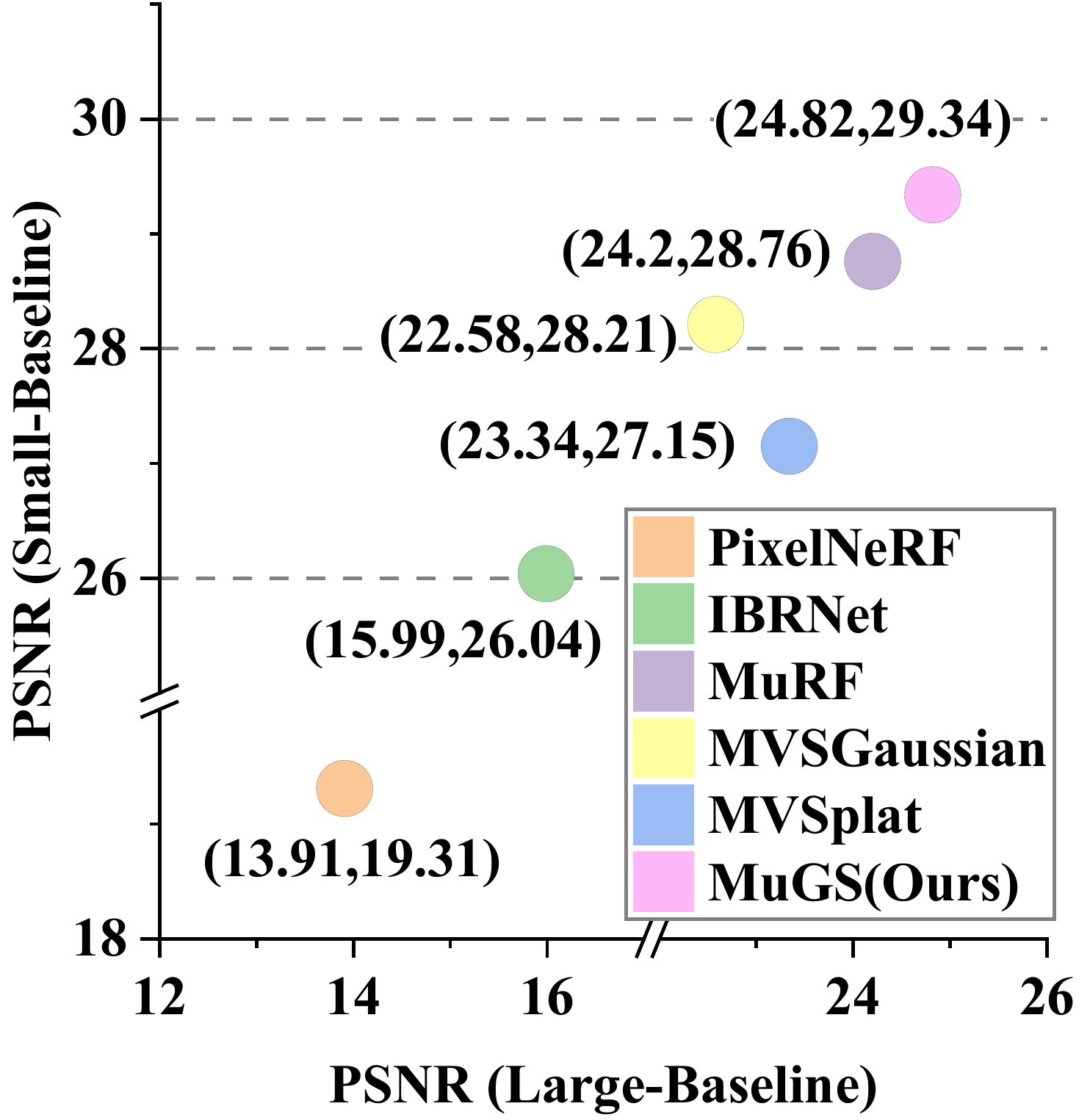}
	\end{minipage} 
	\begin{minipage}[c]{0.515\linewidth}
		\centering
		\includegraphics[width=\textwidth]{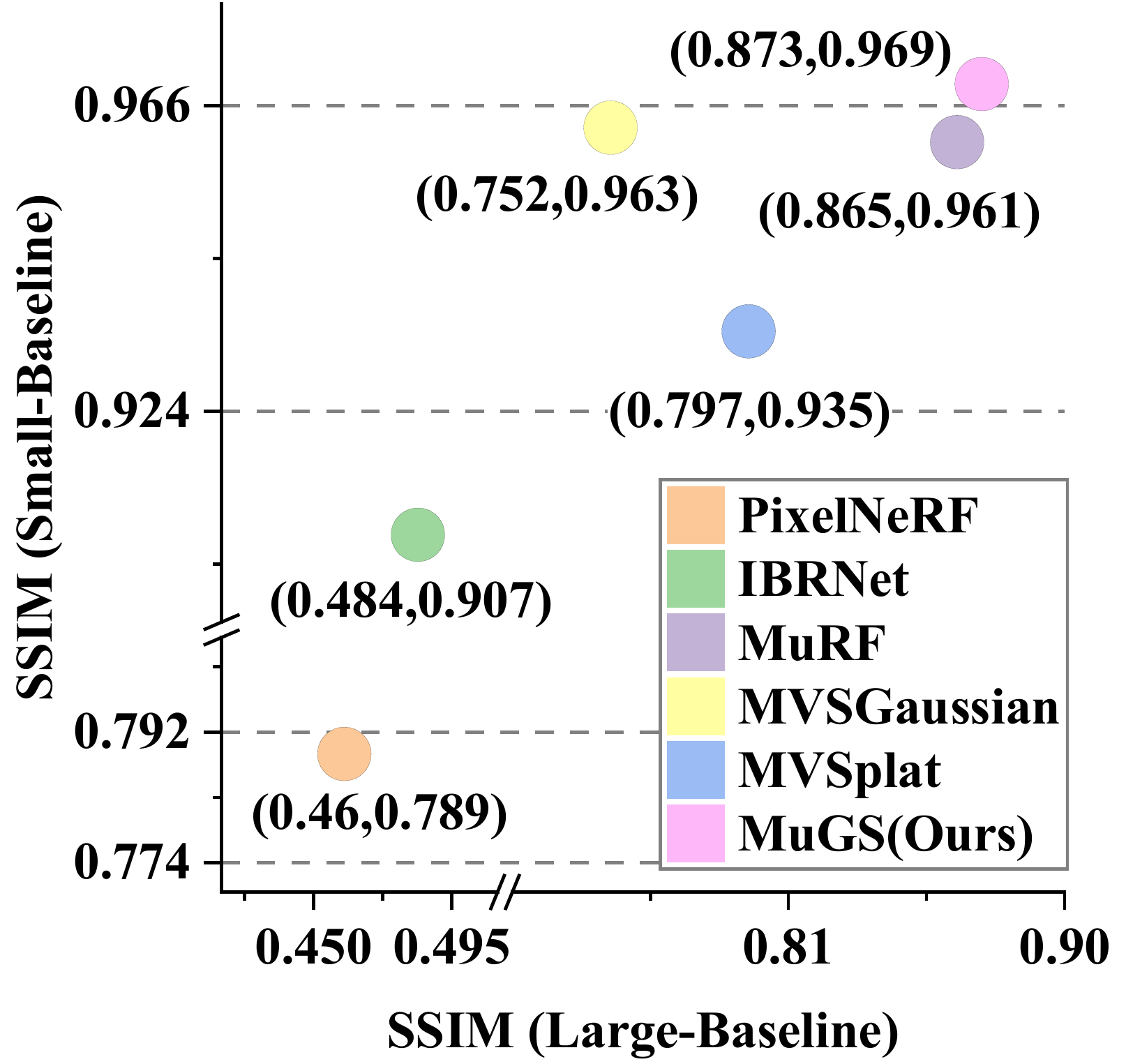}
	\end{minipage} \\
    \subcaption{\nickName\ achieves the best performance in both large- and small-baseline.}
    \end{minipage} \\
    \begin{minipage}[c]{0.8\linewidth}
		\centering
		\includegraphics[width=\textwidth]{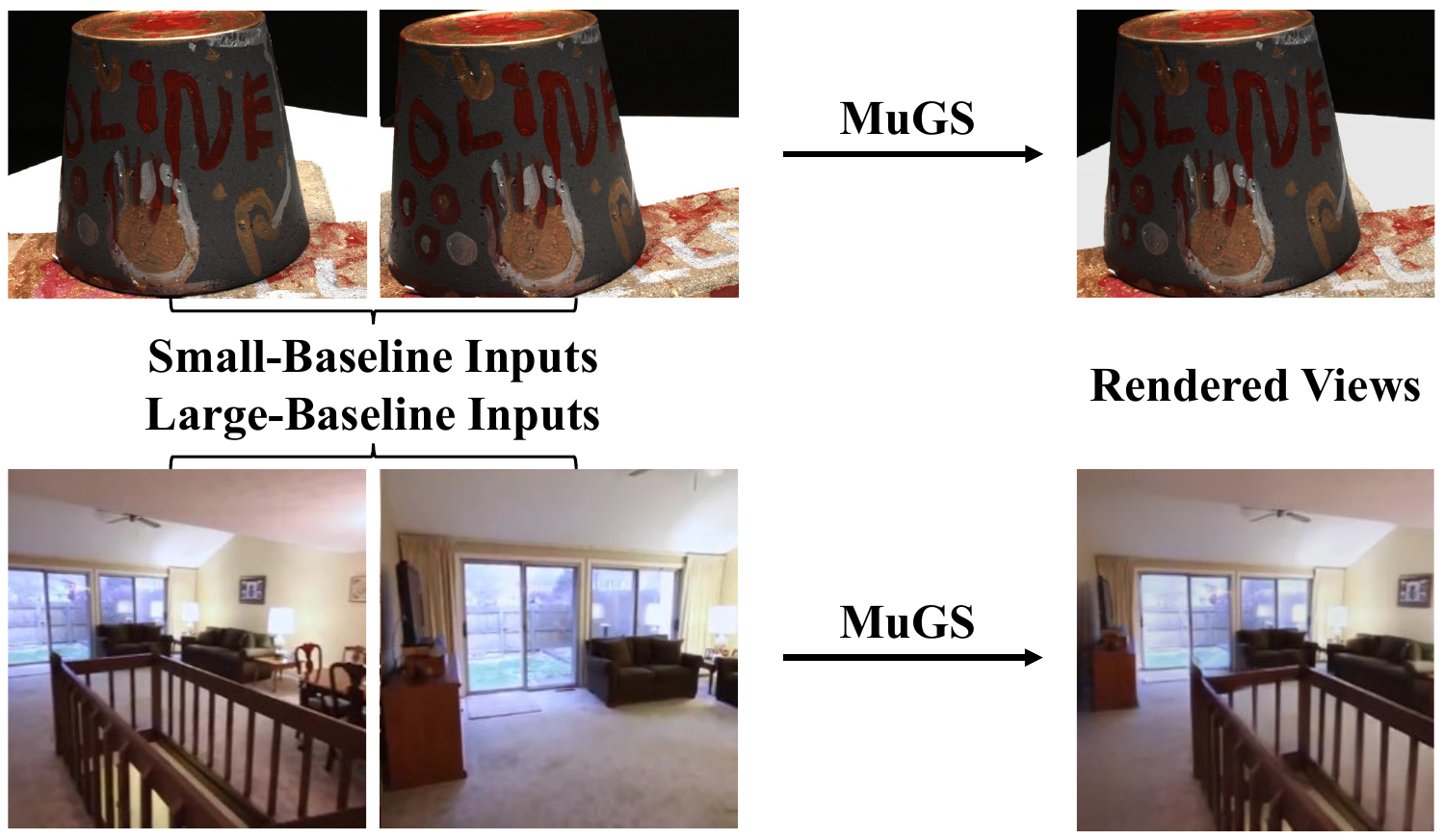}
		\subcaption{\nickName\ can generalize across different baselines.}
    \end{minipage}
    \vspace{-10pt}
	\caption{\textbf{\nickName \ supports multiple baseline settings}. \nickName \ is the first Gaussian-based method designed for different baselines. 
    Our method outperforms the previous state-of-the-art methods.}
	\label{fig:1}
    \vspace{-15pt}
\end{figure}

To tackle this issue, several generalizable methods \cite{liu_mvsgaussian_2024, chen_mvsplat_2024, chen_mvsplat360_2024, charatan_pixelsplat_2024, szymanowicz_splatter_2024, wewer2024latentsplat} achieve notable advancements in rendering high-quality novel views from unseen scenes. These methods accomplish this by introducing explicit geometry constraints and leveraging data-driven approaches rather than overfitting to a specific scene. Based on input view overlap, these methods can be categorized into two classes: small-baseline tasks, which handle images with large overlap, and large-baseline tasks, which operate on images having small overlap. However, existing methods tend to specialize in small-baseline or large-baseline settings, struggling to generalize across different baselines.

A key challenge in practical scenarios is that the baseline of input views are either small or large, which limits the generalization of baseline-specific methods. To address this issue, we propose the first $3$D Gaussian splatting method designed for rendering novel views from sparse inputs across varying baselines, as shown in \cref{fig:1}.

Generalizable Gaussian models typically extract feature volumes and depth probability volumes using multi-view stereo (MVS) techniques \cite{yao_mvsnet_2018, gu_cascade_2020}, then regress depth and other Gaussian parameters. Accurate depth estimation is essential for retrieving reliable information, yet challenges arise with baseline mismatches. Small-baseline models tested on large-baseline datasets suffer from depth errors due to occlusions and insufficient overlap, leading to distortions. Conversely, large-baseline models evaluated on small-baseline datasets struggle with the lack of matching cues, resulting in inaccurate depth estimation. This inaccuracy causes inconsistent Gaussian placements and blurred rendered images.

Our insight is that accurate depth guidance can address common challenges in both large-baseline and small-baseline methods, while unifying them into a more generalized model. However, achieving precise depth under sparse, multi-baseline inputs is challenging due to three key obstacles: \textbf{First}, the preferred depth estimation strategy differs from the baseline. For instance, in a two-view setup, the smallest baseline corresponds to a binocular scenario, where the two views can be treated as adjacent video frames and processed with matching-based MVS techniques. The largest baseline, on the other hand, may imply no overlap between the two views, meaning depth information must rely on monocular depth methods. Resolving both types of problems within a single model is challenging.
\textbf{Second}, unlike typical MVS tasks, we deal with sparse inputs where calculating the feature similarity is often infeasible. This limitation arises because at least two valid feature samples are required for variance at each candidate point. However, sparse inputs may lack sufficient overlap or be affected by occlusions, rendering the process ineffective. 
\textbf{Third}, for effective generalization, the model needs to store comprehensive prior knowledge to support inference across diverse baselines.
Training on a dataset with a specific baseline while preserving multi-baseline adaptability is a nontrivial challenge, requiring careful optimization to prevent overfitting to a particular baseline configuration.

To address the challenges above, we propose the following solutions. \textbf{First}, we introduce a pre-trained monocular depth model \cite{yang_depth_v2_2024} to assist MVS, as the former offers more robust and smooth depth features for sparse inputs, whereas the latter typically exhibits large errors in challenging areas, despite performing well in regions with sufficient context. \textbf{Second}, for each depth candidate in MVS, we compute both projected depth and sampled depth, which represent the spatial position and the expected depth of each point, respectively. A 3D U-net is then employed to calculate the consistency between the two depths. 
\textbf{Third}, the consistency information mentioned above is then used as a query in a lightweight attention network, refining the depth probability volume. By prioritizing depth candidates near the surface, the MLP network better utilizes features and colors sampled from each source view, ultimately reducing artifacts and improving rendering quality. Additionally, we propose a reference-view loss for contextual supervision to learn geometric correspondence more effectively.

Our contributions can be summarized as follows:
\begin{itemize}[leftmargin=*]
\item We propose \nickName, the first multi-baseline generalizable Gaussian based method that integrates the features of multi-view stereo and monocular models.
\item We introduce the projection-sampling depth consistency network to guide the fine-grained probability volume and enhance robustness for challenging sparse inputs.
\item We propose a reference-view loss for contextual supervision to improve rendering quality.
\item We demonstrate that our method outperforms existing approaches across different baseline datasets and achieves superior performance on zero-shot datasets.
\end{itemize}

\section{Related Work}
\label{sec:relate}

\noindent\textbf{Multi-Baseline}. The idea of ``multi-baseline" originates from multi-baseline stereo depth estimation \cite{4587671, 1167864, 1211356, 206955}, in which several stereo pairs with different baselines are employed to overcome matching ambiguities and enhance accuracy. Recently, MuRF \cite{xu_murf_2024} has made significant progress in extending the multi-baseline problem to the NVS task by leveraging a pre-trained multi-view feature encoder to construct target view frustum volume, along with an efficient CNN decoder. This approach can handle both large- and small-baseline problems, even with very sparse inputs. However, as this method relies solely on MVS principles to obtain a density volume, it faces challenges when there is insufficient overlap or occlusion in the views. In such cases, the density along the ray tends to disperse instead of concentrating around the true surface. Moreover, due to the NeRF-like volume rendering approach, noise feature sampled from incorrect depths also contributes to the final output, resulting in blurriness and artifacts. In contrast, our work addresses these challenges from a more fundamental perspective, specifically depth precision. By doing so, we propose a unified solution that effectively handles the shared challenges encountered by both large-baseline and small-baseline methods.

\noindent\textbf{Multi-View Stereo (MVS)} aims to recover 3D geometry from multiple views. Traditional MVS methods \cite{fua1995object, galliani2015massively, schonberger2016pixelwise, schonberger2016structure} rely on handcrafted features and similarity metrics, limiting performance, while MVSNet \cite{yao_mvsnet_2018} first introduces an end-to-end pipeline that constructs a cost volume to aggregate 2D data in a 3D geometry-aware manner.
Following this cost volume-based pipeline, subsequent works make improvements from various aspects, \eg higher memory efficiency with coarse-to-fine architectures \cite{gu_cascade_2020, cheng2020deep, yan2020dense} or recurrent plane sweeping \cite{yan2020dense, yao2019recurrent}, optimized cost aggregation \cite{wang2022mvster, wei2021aa}, enhanced feature representations \cite{liu2023epipolar, ding2022transmvsnet}, and improved decoding strategy \cite{peng2022rethinking, ye2023constraining}. As the cost volume encodes the consistency of multi-view features and naturally performs correspondence matching, many feed-forward Gaussian methods \cite{chen_mvsplat_2024, liu_mvsgaussian_2024, chen_mvsplat360_2024} follow this spirit to learn better geometry. However, they inherently suffer from the limitation of feature matching in challenging situations like insufficient overlap or occlusion.

\noindent\textbf{Monocular Depth Estimation (MDE)}. Recently, there has been notable advancement in depth estimation from a single image \cite{ranftl2020towards, bhat2023zoedepth, yin2023metric3d, ke2024repurposing, yang_depth_2024}, with current methods delivering impressively accurate edge-aligned results across a wide range of real-world data. However, monocular depth techniques still face challenges with scale ambiguities and are unable to generate depth predictions that are consistent across multiple views, which are essential for tasks like 3D reconstruction \cite{yin2022towards} and video depth estimation \cite{wang2023neural}. In this paper, we propose the concepts of projected depth and sampled depth to integrate depth information from both the cost volume and monocular depth. By leveraging the robustness of a pre-trained monocular depth model \cite{yang_depth_v2_2024}, our approach mitigates the limitations of feature matching-based methods. The very recent work DepthSplat ~\cite{xu2024depthsplat} attempts to combine MVS and MDE, focusing on the mutual enhancement of depth estimation and large-baseline view synthesis through a feature-level concatenation. In contrast, our work deeply explores and models the relationship between the two depth cues across different view baselines, enabling view synthesis under varying baselines.
\section{Preliminary}
\label{sec:pre}

\noindent\textbf{Adjusted Multi-View Stereo Pipeline} adapts the traditional MVS approach for novel view synthesis (NVS). The process starts by defining multiple fronto-parallel planes in the target view. Then, the feature maps extracted from \(N\) input views are warped onto these planes using a differentiable homography, expressed as:
\begin{equation}
    H_i(z)=K_t R_t \mathbb{C}(z) R_i^{-1} K_i^{-1},
    \label{eq:homo}
\end{equation}
where \([K_t, R_t]\) and \([K_i, R_i]\) are the camera intrinsics and rotations for the target view and the source view \(I_i\). With \(\mathbb{I}\) the identity matrix, \(z\) the depth candidate, \(n\) the principal axis of the target view camera and \([t_t,t_i]\) the camera translations for the target view and the source view \(I_i\), we can obtain the correction term \(\mathbb{C}(z)\) by:
\begin{equation}
    \mathbb{C}(z)=\mathbb{I} - \frac{(R_t^{-1} t_t - R_i^{-1} t_i) n^T R_t}{z}.
\end{equation}

With warped multi-view features \(\{f_i\}_{i=1}^N\), we can obtain the cost volume by calculating learnable pair-wise similarity \cite{zhang2023vis}. It can be expressed as:
\begin{equation}
    Sim=\sum_{j<k} w_{jk} * cos(f_j, f_k), \quad j,k\in\{1,2,...,N\},
\end{equation}
where \(w_{jk}\) are learned weights.

For novel view synthesis, it is essential not only to focus on depth but also to recover textures and colors. Thus, the cost volume is augmented by introducing multi-view features which are aggregated through a pooling network \cite{lin2022efficient}. The augmented cost volume is regularized by CNNs to produce the target view frustum volume, from which we obtain the depth probability volume \(\mathbb{V}^p\) and other rendering parameters via MLPs. Unlike previous studies \cite{liu_mvsgaussian_2024, xu_murf_2024}, which directly use the depth probability volume for rendering, our approach improves both depth and texture recovery, enhancing the final rendering quality.

\noindent\textbf{3D Gaussian Splatting} uses anisotropic 3D Gaussians to explicitly represent a 3D scene. Each Gaussian is defined by:
\begin{equation}
    G(X)=exp[-\frac{1}{2}(X-\mu)^T \Sigma^{-1}(X-\mu)],
\end{equation}
where \(\Sigma\) and \(\mu\) denotes 3D covariance matrix and mean. The covariance matrix \(\Sigma\) is usually decomposed into a scaling matrix \(S\) and a rotation matrix \(R\) by \(\Sigma=RSS^TR^T\), which allows for effective optimization since \(\Sigma\) holds physical meaning only when it is positive semidefinite.

To render a view from 3D Gaussians, the first step is to splat Gaussians from 3D space to a 2D plane, yielding 2D Gaussians, which covariance matrix is calculated by \(\Sigma'=JW\Sigma W^TJ^T\). The \(J\) is the Jacobian matrix which represents the affine approximation of the projective transformation, and the \(W\) is the view transformation matrix. Next, the color of each pixel can be rendered by alpha-blending:
\begin{equation}
    C=\sum\nolimits_j c_j\alpha_j \prod\nolimits_{k=1}^{j-1}(1-\alpha_k),
\end{equation}
where the color \(c_j\) at depth-wise position \(j\) is defined by spherical harmonics (SH) coefficients and the density \(\alpha_j\) equals to the multiplication of 2D Gaussians and a learnable point-wise opacity.
\section{Methodology}
\label{sec:method}

\begin{figure*}[htbp]
\centering
\includegraphics[width=1.0\linewidth]{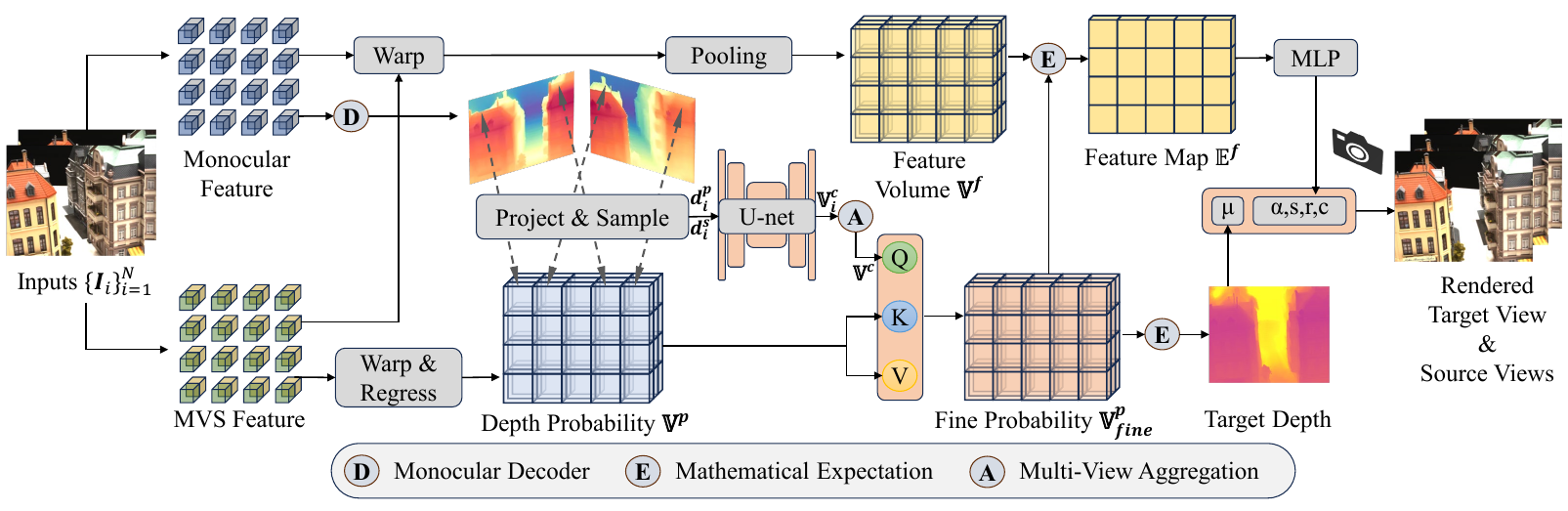}
\vspace{-20pt}
\caption{\textbf{Overview.} Given input images \({\{I_i\}}_{i=1}^N\), we first extract multi-view image features from both the monocular encoder and MVS's cross-view encoder. The MVS features are used to regress a target view depth probability volume, while monocular features are decoded into source view depth maps \({\{\mathcal{D}_i\}}_{i=1}^N\). By projecting the points of the depth probability volume to and sampling from the depth map \(\mathcal{D}_i\), we obtain \(d^p_i\) and \(d^s_i\), which are then fed into a U-net to query for a refined probability volume \(\mathbb{V}^p_{fine}\). Besides, both features are concatenated to construct the feature volume. Next, we calculate the expected value of depth and feature using \(\mathbb{V}^p_{fine}\), which produces the target depth and feature map. These are used to predict Gaussian parameters. Finally, the target view image and source reference views are rendered, which contribute to the total loss together.}
\label{fig:main}
\vspace{-15pt}
\end{figure*}

\subsection{Overview}
Given a set of input views \({\{I_i\}}_{i=1}^N\), our objective is to render target views through a feed-forward, generalized process without per-scene optimization. The overview of our proposed framework is depicted in \cref{fig:main}. Our method consists of two primary branches: the MVS branch and the MDE branch. In the MVS branch, a multi-view feature encoder is applied to the input views, constructing the target view frustum volume to regress a coarse depth probability volume. Meanwhile, the MDE branch generates monocular feature maps and predicts monocular depth maps for input views. Subsequently, using a projection-and-sampling approach, monocular depth information from multiple views is integrated with the depth probability volume, refining the predicted target view depth. On the other hand, MVS features are enhances by monocular features to obtain the feature volume, which is then retrieved for the feature map by the refined probability volume, enabling the MLPs to regress Gaussian parameters while reducing noise. This pipeline is executed hierarchically, generating depth maps and rendered views in a coarse-to-fine manner.

\subsection{MDE-based Depth Refining}
\noindent\textbf{Fusion Strategy}. The most straightforward approach to fusing monocular depth estimation (MDE) and multi-view stereo (MVS) depth would be to merge both target view aligned monocular depth map and target view MVS depth probability volume together through a neural network \cite{li2023learning, yang2022mvs2d}. However, this method proves inadequate for our novel view synthesis task, as it faces a fundamental limitation: while MVS can estimate the depth map for the target view, the corresponding monocular depth map is inherently unavailable since the target view itself is what we aim to generate. Therefore, we propose a fusion strategy based on projection and sampling manner. Specifically, after obtaining the depth probability volume, we first estimate the monocular depth map for each input view, \ie, \(\{\mathcal{D}_i\}_{i=1}^N\). We then calculate projected and sampled depth information. Given the camera intrinsic \(K_i\), rotation \(R_i\) and translation \(t_i\) of the input view \(I_i\), each depth candidate point \(\mathcal{P}\) on the fronto-parallel plane can be projected to input view \(I_i\) by:
\begin{equation}
    \mathcal{P}_i*d^p_i=K_i(R_i \mathcal{P} + t_i),
\end{equation}
where we can simultaneously obtain both the the projected depth \(d^p_i\), which is the distance from the point to the camera plane, and the projection coordinates \(\mathcal{P}_i\) in the camera coordinate system.
The sampled depth \(d^s_i\) is obtained by performing grid sampling on the monocular depth map \(\mathcal{D}_i\) according to the projection coordinates \(\mathcal{P}_i\).

For candidate points near the object's surface, the projected depth and sampled depth exhibit a high degree of consistency, as both represent the same spatial location. In contrast, for candidates far from the object surface, the two depths become inconsistent. This property enables us to infer the authenticity of a candidate point based on the consistency between the two depths. To leverage this insight, we subsequently employ a four-layer 3D U-Net \(\mathcal{U}\) for each view \(i\) to regress consistency cue \(\mathbb{V}^c_i\) from the volume composed of \(d^p_i\) and \(d^s_i\). Additionally, to mitigate the inherent scale ambiguity in monocular depth estimation, we introduce the depth ratio \(d^s_i/d^p_i\) as a third input channel. This normalization helps reduce discrepancies caused by varying depth scales across different views.
\begin{equation}
    \mathbb{V}_i^c=\mathcal{U}(d^p_i, d^s_i, d^s_i/d^p_i).
\end{equation}
Each consistency cue \(\mathbb{V}^c_i\) indicates the consistency between the MVS's target view depth probability volume and the MDE's source view depth map \(\mathcal{D}_i\). This operation is executed on every input view, yielding the set of the consistency cues \({\{\mathbb{V}^c_i\}}_{i=1}^N\).

\noindent\textbf{Probability Refinement}. With the multi-view consistency cues \({\{\mathbb{V}^c_i\}}_{i=1}^N\), we first aggregate them based on the visibility, which can be expressed as:
\begin{equation}
    \mathbb{V}^c=\sum\nolimits_{i=1}^N w_i \mathbb{V}^c_i,
\end{equation}
where \(w_i\) are learnable weights and the outcome \(\mathbb{V}^c\) serves as the overall consistency cue. This aggregation allows the model to focus on those informative cues and discard the noisy ones caused by occlusion.

To refine the depth probability volume, we not only consider the consistency between projected and sampled depths, but also take into account the original MVS estimation results, since the latter cab be credible if sufficient context is given. Therefore, we use a lightweight attention network to integrate consistency cues with MVS results, which helps the network to balance the information. Specifically, we take consistency cue \(\mathbb{V}^c\) as the query, the depth probability volume \(\mathbb{V}^p\) as both the key and value, to conduct a depth-wise attention. 
The output result has the dimension of depth probability and is added with the residual of the original volume, which can be expressed as:
\begin{equation}
    \mathbb{V}^p_{fine}=Attention(\mathbb{V}^c, \mathbb{V}^p,\mathbb{V}^p) + \mathbb{V}^p
\end{equation}

\subsection{Gaussian Parameter Prediction}
\noindent\textbf{Feature Enhancement}. The import of a monocular model provides not only depth information but also well-encoded features, which carry informative inductive bias. To this end, we leverage the power of the monocular features to assist the prediction of Gaussian parameters. Specifically, both the monocular feature maps and the MVS feature maps are warped to the target view fronto-parallel planes according to \cref{eq:homo} and concatenated. Subsequently, utilizing a pooling network~\cite{lin2022efficient}, the features from different views are aggregated to construct the feature volume \(\mathbb{V}^f\).

\noindent\textbf{Gaussian Construction}. With the refined depth probability volume, we can first compute the expected values of depths, yielding the target view depth map. The depths are then used to unproject each pixel to obtain the positions \(\mu\) of the 3D Gaussians. Next, we regress from feature volume \(\mathbb{V}^f\) to obtain the remaining parameters for each Gaussian, namely the color \(c\), opacity \(\alpha\), scale \(s\), and rotation \(r\). Unlike NeRF-based methods that require the whole volume to be processed and prone to vaporific noise in the result, our method is constructed on a 2D feature map and focuses more on features describing the actual object's surface. Specifically, we compute the expectation of the feature volume along the depth channel using the refined depth probability, and the resulting features \(\mathbb{E}^f\) are used as the input to the MLP \(\phi\) to calculate:
\begin{equation}
    \begin{split}
        c &=Sigmoid(\phi_c(\mathbb{E}^f)), \alpha=Sigmoid(\phi_{\alpha}(\mathbb{E}^f)), \\
        s &=Softplus(\phi_s(\mathbb{E}^f)), r=Norm(\phi_r(\mathbb{E}^f)).   
    \end{split}
\end{equation}
With the set \(\{\mu,s,r,\alpha,c\}\), pixel-aligned Gaussians are represented and alpha-blending can be performed to obtain the color of each pixel.

\subsection{Multi-View Gaussian Splatting}
\label{subsec:mvgs}
Due to the characteristic of target view pixel-aligned Gaussians, using only target view RGB supervision during training limits the ability to reflect the spatial information of the Gaussians in the results, ultimately hindering the achievement of accurate geometry. The explicit Gaussian representation allows us to quickly render not only the novel view but also source views without additionally constructing source view volumes. Therefore, we incorporate supervision from the source views to improve spatial accuracy. Specifically, after obtaining the parameter set \(\{\mu,s,r,\alpha,c\}\), we input the camera parameters of both the source and target views, \ie \(\{[K_i,R_i]\}_{i=1}^N\) and \([K_t,R_t]\) into the Gaussian rasterizer to generate multiple rendered views in sequence for optimization rapidly.

\subsection{Training Objective}
\noindent\textbf{Hierarchical Training}. Our model is trained level-by-level in a coarse-to-fine manner. Specifically, the depth probability volume constructed by the previous level is transformed into a probability distribution function (PDF), which is then utilized to sample a smaller number of more accurate depth candidates for the subsequent level. This approach facilitates a more precise and memory-efficient training and rendering process.

\noindent\textbf{Training Loss}. 
Our model is trained solely under the supervision of RGB images. Different from the existing feed-forward 3D-GS methods \cite{liu_mvsgaussian_2024, chen_mvsplat_2024, charatan_pixelsplat_2024}, we introduce a novel approach by integrating reference views into the overall supervision as the reference loss $\mathcal{L}^{src}$. The inclusion of additional contextual views not only enhances geometric cues but also enriches texture information, thereby accelerating optimization and improving rendering quality. 
Specifically, for each layer \(k\), the loss function $\mathcal{L}_{total}^k$ comprises both the target view loss and the source view loss. The former includes L1 loss, SSIM loss \cite{wang2004image}, and perceptual loss \cite{zhang2018unreasonable}, while the latter consists of L1 loss computed for each individual source view, as demonstrated in \cref{subsec:mvgs}. The overall loss can be computed by:
\begin{equation}
    \mathcal{L}_{total}^k = \mathcal{L}_1^{target} + \mathcal{L}_{SSIM} + \mathcal{L}_{LPIPS} + \sum\nolimits_{i=1}^N {\mathcal{L}_1^{src}}_i.
\end{equation}
\section{Experiments}
\label{sec:exper}

\subsection{Settings}
\begin{table*}[htbp]
\begin{minipage}[t]{0.67\linewidth}
  \centering
  \caption{DTU small-baseline.}
  \vspace{-10pt}
    \resizebox{\linewidth}{!}
    {\begin{tabular}{lcccccccccc}
    \toprule
    \multirow{2}[4]{*}{Method} & \multicolumn{3}{c}{3-view} &       & \multicolumn{3}{c}{2-view} &       & \multicolumn{2}{c}{Inference Time (s)} \\
\cmidrule{2-4}\cmidrule{6-8}\cmidrule{10-11}          & PSNR↑ & SSIM↑ & LPIPS↓ &       & PSNR↑ & SSIM↑ & LPIPS↓ &       & Encode↓ & Render↓ \\
    \midrule
    PixelNeRF~\cite{yu2021pixelnerf} & 19.31 & 0.789 & 0.382 &       & -     & -     & -     &       & \textbf{0.005} & 5.294 \\
    IBRNet~\cite{wang2021ibrnet} & 26.04 & 0.907 & 0.191 &       & -     & -     & -     &       & \underline{0.016} & 4.592 \\
    MVSNeRF~\cite{chen_mvsnerf_2021} & 26.63 & 0.931 & 0.168 &       & 24.03 & 0.914 & 0.192 &       & 0.042 & 2.363 \\
    ENeRF~\cite{lin2022efficient} & 27.61 & 0.957 & 0.089 &       & 25.48 & 0.942 & 0.107 &       & 0.019 & 0.032 \\
    MuRF~\cite{xu_murf_2024}  & \underline{28.76} & 0.961 & 0.077 &       & \underline{27.02} & \underline{0.949} & \underline{0.088} &       & 0.142 & 1.122 \\
    \midrule
    PixelSplat~\cite{charatan_pixelsplat_2024} & -     & -     & -     &       & 14.01 & 0.662 & 0.389 &       & 0.102 & \underline{2.3E-3} \\
    MVSplat~\cite{chen_mvsplat_2024} & 27.15 & 0.935 & 0.121 &       & 25.02 & 0.915 & 0.126 &       & 0.040 & 3.9E-3 \\
    MVSGaussian~\cite{liu_mvsgaussian_2024} & 28.21 & \underline{0.963} & \underline{0.076} &       & 25.78 & 0.947 & 0.095 &       & 0.021 & 2.4E-3 \\
    Ours  & \textbf{29.34} & \textbf{0.969} & \textbf{0.075} &       & \textbf{27.56} & \textbf{0.958} & \textbf{0.084} &       & 0.153 & \textbf{2.1E-3} \\
    \bottomrule
    \end{tabular}%
    }
  \label{tab:dtu}%
\end{minipage}
\begin{minipage}[t]{0.325\linewidth}
  \centering
  \caption{RealEstate10K large-baseline.}
  \vspace{-10pt}
  \resizebox{\linewidth}{!}
    {\begin{tabular}{lccc}
    \toprule
    Method & PSNR↑ & SSIM↑ & LPIPS↓ \\
    \midrule
    PixelNeRF~\cite{yu2021pixelnerf} & 13.91 & 0.46  & 0.591 \\
    SRF~\cite{chibane2021stereo}   & 15.40 & 0.486 & 0.604 \\
    GeoNeRF~\cite{johari_geonerf_2022} & 16.65 & 0.511 & 0.541 \\
    IBRNet~\cite{wang2021ibrnet} & 15.99 & 0.484 & 0.532 \\
    GPNR~\cite{suhail2022generalizable}  & 18.55 & 0.748 & 0.459 \\
    AttnRend~\cite{du2023learning} & 21.38 & 0.839 & 0.262 \\
    MuRF~\cite{xu_murf_2024}  & \underline{24.20} & \underline{0.865} & \underline{0.170} \\
    \midrule
    MVSGaussian~\cite{liu_mvsgaussian_2024} & 22.58 & 0.752 & 0.206 \\
    MVSplat~\cite{chen_mvsplat_2024} & 23.34 & 0.797 & 0.188 \\
    MuGS(Ours)  & \textbf{24.82} & \textbf{0.873} & \textbf{0.153} \\
    \bottomrule
    \end{tabular}%
    }
  \label{tab:re10k}%
\end{minipage}
\vspace{-20pt}
\end{table*}%

\noindent\textbf{Datasets}. To evaluate the cross-baseline performance of our method, we select two widely used datasets for training and testing: the object-centric dataset DTU \cite{jensen_large_2014}, which can provide small-baseline inputs and the RealEstate10K \cite{zhou_stereo_2018} which can serve as the large-baseline dataset. To further evaluate the generalizable performance, we select the forward-facing dataset LLFF \cite{mildenhall_nerf_2020} and the large-baseline Mip-NeRF 360 dataset \cite{barron_mip-nerf_2022} to conduct zero-shot evaluation.

\noindent\textbf{Baselines}. We compare our method against several generalizable NeRF-based methods \cite{yu2021pixelnerf, wang2021ibrnet, chen_mvsnerf_2021, lin2022efficient, johari_geonerf_2022, suhail2022generalizable, du2023learning} as well as two typical 3D-GS methods MVSplat \cite{chen_mvsplat_2024} and MVSGaussian \cite{liu_mvsgaussian_2024} which are designed for large-baseline inputs and small-baseline inputs, respectively. Besides, we compare our method with MuRF \cite{xu_murf_2024}, the state-of-the-art NeRF-based multi-baseline method.

\noindent\textbf{Implementation Details}. Our implementation is mainly based on PyTorch and the 3D-GS rendering implemented in CUDA. We sample 64 isometric depth candidates for the coarse model and 16 for the fine model. All models are trained on 2 Nvidia A6000 GPUs with the Adam \cite{kinga2015method} optimizer. For the pre-trained monocular model, we use Depth Anything V2 \cite{yang_depth_v2_2024}. 

\begin{figure}[t]
\centering
\includegraphics[width=1.0\linewidth]{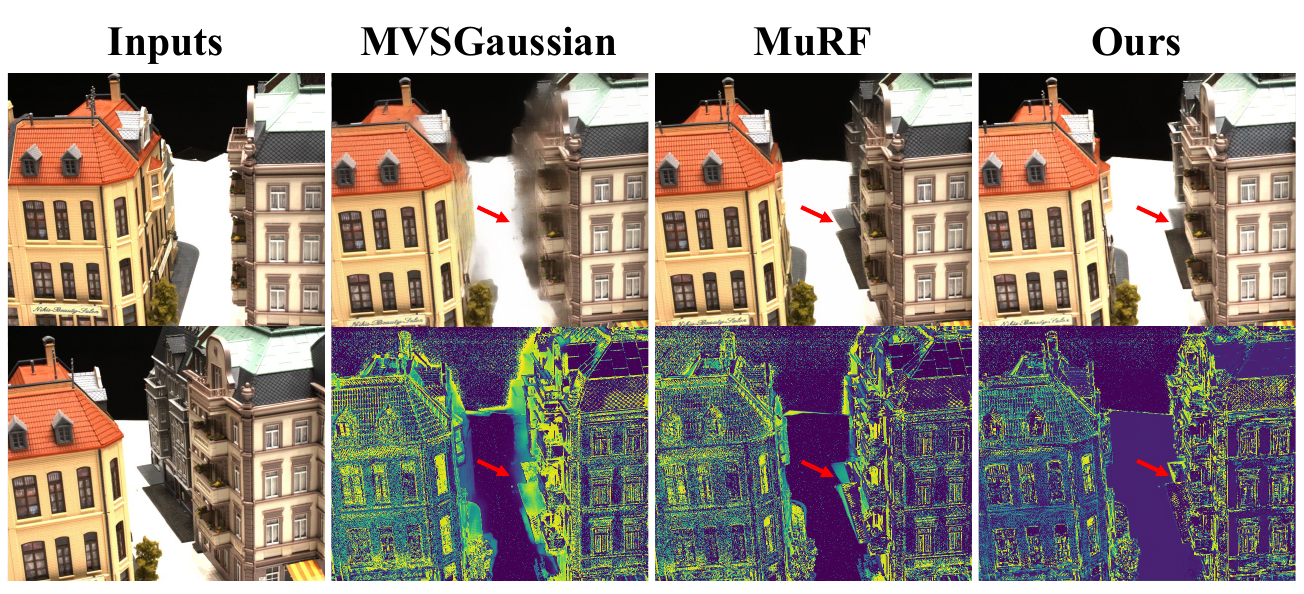}
\vspace{-20pt}
\caption{\textbf{2-view small-baseline results on the DTU~\cite{jensen_large_2014} dataset}. Our method renders higher quality with fewer errors than other small-baseline methods.}
\label{fig:dtu}
\vspace{-15pt}
\end{figure}

\subsection{Results}
\noindent\textbf{Small-Baseline on DTU}. The DTU dataset \cite{jensen_large_2014} is a small-baseline dataset since the input images are object-centric and provide significant overlap between views. We follow the setting of MuRF \cite{xu_murf_2024}, which takes the nearest 3 views around the target view to serve as source views. Additionally, we evaluate the 2-view scenario, which is more challenging since there is more occlusion and less context. 
We achieve more than 0.5dB PSNR improvement compared to the previous best methods in both 2-view and 3-view settings. Besides, our method provides higher inference speed compared to MuRF thanks to the Gaussian representation.

As shown in \cref{fig:dtu}, this setting presents a significant challenge due to the large occlusion between the \textit{buildings} and the limited availability of only two input views. Since the texture in the occluded regions is visible in only one of the input views, methods such as MVSGaussian \cite{liu_mvsgaussian_2024}, which heavily rely on MVS pipeline, struggle to predict accurate depth. This results in blurry rendered output images. While MuRF \cite{xu_murf_2024} achieves better quality than MVSGaussian, it still exhibits geometric inaccuracies, as the marked area is misplaced in \cref{fig:dtu}. In contrast, our method demonstrates superior performance in both rendering quality and geometric accuracy, which we attribute this performance to our fusion strategy and multi-view training supervision.

\begin{figure}[t]
\centering
\includegraphics[width=1.0\linewidth]{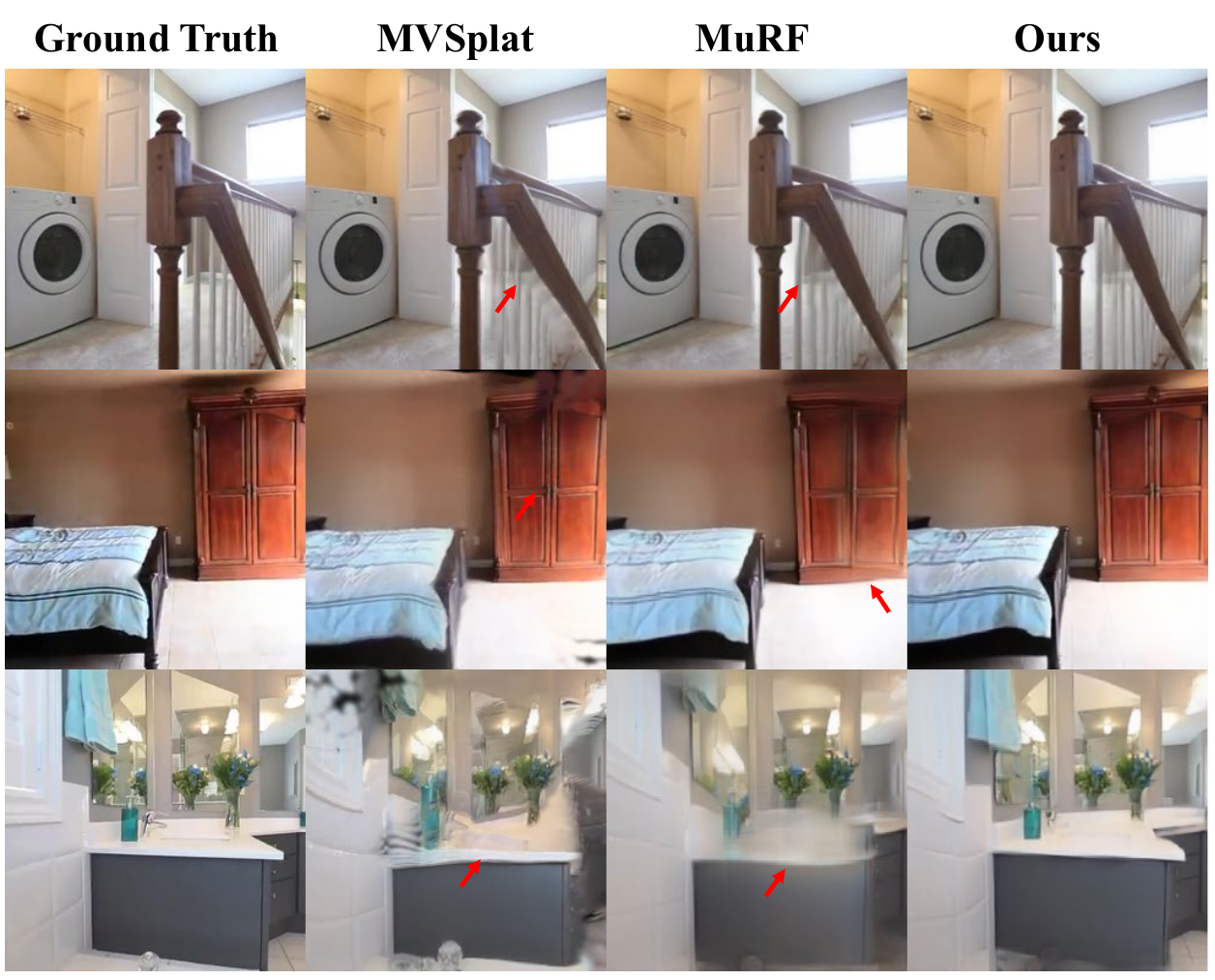}
\vspace{-20pt}
\caption{\textbf{2-view large-baseline results on the RealEstate10K dataset}. The images rendered by our method exhibit superior geometric accuracy and reduced distortion.}
\label{fig:re10k}
\vspace{-15pt}
\end{figure}
\begin{table*}[thbp]
\begin{minipage}[t]{0.52\linewidth}
\begin{minipage}[t]{\linewidth}
  \centering
  \caption{Depth evaluation results from 3-view inputs on DTU.}
  \vspace{-10pt}
  \resizebox{\linewidth}{!}
    {\begin{tabular}{lccccccc}
    \toprule
    \multirow{2}[4]{*}{Method} & \multicolumn{3}{c}{Reference view} &       & \multicolumn{3}{c}{Novel view} \\
\cmidrule{2-4}\cmidrule{6-8}          & Abs err↓ & Acc(2)↑ & Acc(10)↑ &       & Abs err↓ & Acc(2)↑ & Acc(10)↑ \\
    \midrule
    MVSNet~\cite{yao_mvsnet_2018} & 3.60  & 0.603 & 0.955 &       & -     & -     & - \\
    PixelNeRF~\cite{yu2021pixelnerf} & 49.0  & 0.037 & 0.176 &       & 47.8  & 0.039 & 0.187 \\
    IBRNet~\cite{wang2021ibrnet} & 338   & 0.000 & 0.913 &       & 324   & 0.000 & 0.866 \\
    MVSNeRF~\cite{chen_mvsnerf_2021} & 4.60  & 0.746 & 0.913 &       & 7.00  & 0.717 & 0.866 \\
    ENeRF~\cite{lin2022efficient} & 3.80  & 0.837 & 0.939 &       & 4.60  & 0.792 & 0.917 \\
    MuRF~\cite{xu_murf_2024}  & -     & -     & -     &       & 12.73 & 0.583 & 0.906 \\
    MVSGaussian~\cite{liu_mvsgaussian_2024} & \textbf{3.11} & \underline{0.866} & \underline{0.956} &       & \underline{3.66}  & \underline{0.838} & \underline{0.945} \\
    MuGS(Ours)  & \underline{3.23}  & \textbf{0.872} & \textbf{0.963} &       & \textbf{3.52} & \textbf{0.853} & \textbf{0.952} \\
    \bottomrule
    \end{tabular}%
    }
  \label{tab:depth}%
\end{minipage}

\begin{minipage}[t]{\linewidth}
    \centering
  \caption{Zero-shot performance on DTU and Mip-NeRF 360 dataset.}
  \vspace{-10pt}
    \resizebox{\linewidth}{!}
    {\begin{tabular}{lccccccc}
    \toprule
    \multirow{2}[4]{*}{Method} & \multicolumn{3}{c}{DTU} &       & \multicolumn{3}{c}{Mip-NeRF 360 Dataset} \\
\cmidrule{2-4}\cmidrule{6-8}          & PSNR↑ & SSIM↑ & LPIPS↓ &       & PSNR↑ & SSIM↑ & LPIPS↓ \\
    \midrule
    AttnRend~\cite{du2023learning} & 11.35 & 0.567 & 0.651 &       & 14.00 & 0.474 & 0.712 \\
    MVSplat~\cite{chen_mvsplat_2024} & 13.94 & 0.473 & 0.385 &       & -     & -     & - \\
    MVSGaussian~\cite{liu_mvsgaussian_2024} & 19.26 & 0.716 & 0.284 &       & 21.19 & 0.752 & 0.322 \\
    MuRF~\cite{xu_murf_2024}  & \underline{22.19} & \underline{0.894} & \underline{0.211} &       & \underline{23.98} & \underline{0.800} & \underline{0.293} \\
    MuGS(Ours)  & \textbf{22.43} & \textbf{0.916} & \textbf{0.202} &       & \textbf{24.25} & \textbf{0.845} & \textbf{0.256} \\
    \bottomrule
    \end{tabular}%
    }
  \label{tab:mip}%
\end{minipage}
\end{minipage}
\begin{minipage}[t]{0.475\linewidth}
  \centering
  \caption{Zero-shot performance on LLFF after trained in DTU.}
  \vspace{-10pt}
    \resizebox{\linewidth}{!}
    {\begin{tabular}{lcccc}
    \toprule
    Method & \multicolumn{1}{l}{Settings} & PSNR↑ & SSIM↑ & LPIPS↓ \\
    \midrule
    PixelNeRF~\cite{yu2021pixelnerf} & \multirow{8}[2]{*}{3-view} & 11.24 & 0.486 & 0.671 \\
    IBRNet~\cite{wang2021ibrnet} &       & 21.79 & 0.786 & 0.279 \\
    MVSNeRF~\cite{chen_mvsnerf_2021} &       & 21.93 & 0.795 & 0.252 \\
    ENeRF~\cite{lin2022efficient} &       & 23.63 & 0.843 & 0.182 \\
    MatchNeRF~\cite{chen2023explicit} &       & 22.43 & 0.805 & 0.244 \\
    MuRF~\cite{xu_murf_2024}  &       & 23.67 & \underline{0.860} & 0.206 \\
    MVSGaussian~\cite{liu_mvsgaussian_2024} &       & \underline{24.07} & 0.857 & \textbf{0.164} \\
    MuGS(Ours)  &       & \textbf{24.21} & \textbf{0.872} & \underline{0.165} \\
    \midrule
    MVSNeRF~\cite{chen_mvsnerf_2021} & \multirow{6}[2]{*}{2-view} & 20.22 & 0.763 & 0.287 \\
    ENeRF~\cite{lin2022efficient} &       & 22.78 & 0.821 & 0.191 \\
    MatchNeRF~\cite{chen2023explicit} &       & 20.59 & 0.775 & 0.276 \\
    MuRF~\cite{xu_murf_2024}  &       & 22.82 & \underline{0.846} & 0.208 \\
    MVSGaussian~\cite{liu_mvsgaussian_2024} &       & \underline{23.11} & 0.834 & \underline{0.175} \\
    MuGS(Ours)  &       & \textbf{23.33} & \textbf{0.855} & \textbf{0.169} \\
    \bottomrule
    \end{tabular}%
    }
  \label{tab:llff}%
\end{minipage}
\vspace{-20pt}
\end{table*}%
\noindent\textbf{Large-Baseline on RealEstate10K}. In this dataset~\cite{zhou_stereo_2018}, we follow the setting of AttnRend \cite{du2023learning} and MuRF \cite{xu_murf_2024}, where the 2 input views are selected from a video with a distance of 128 frames, and the target view to synthesis is an intermediate frame. This large-baseline setting provides relatively small overlap, which is challenging for methods like MVSGaussian \cite{liu_mvsgaussian_2024}, which relies heavily on multi-view feature matching, and monocular cues can be useful to inference geometry.
As shown in \cref{tab:re10k}, our method achieves more than $1$dB PSNR improvement compared to the previous best large-baseline 3D-GS method MVSplat as well as more than $0.5$dB PSNR improvement compared to the previous state-of-the-art multi-baseline method MuRF. The visual comparison in \cref{fig:re10k} indicates that our method produces clearer rendering results than MVSplat \cite{chen_mvsplat_2024} and MuRF \cite{xu_murf_2024}. Meanwhile, the images generated by our method show both geometrical precision and reduced distortion.

\begin{figure}[t]
\centering
\includegraphics[width=1.0\linewidth]{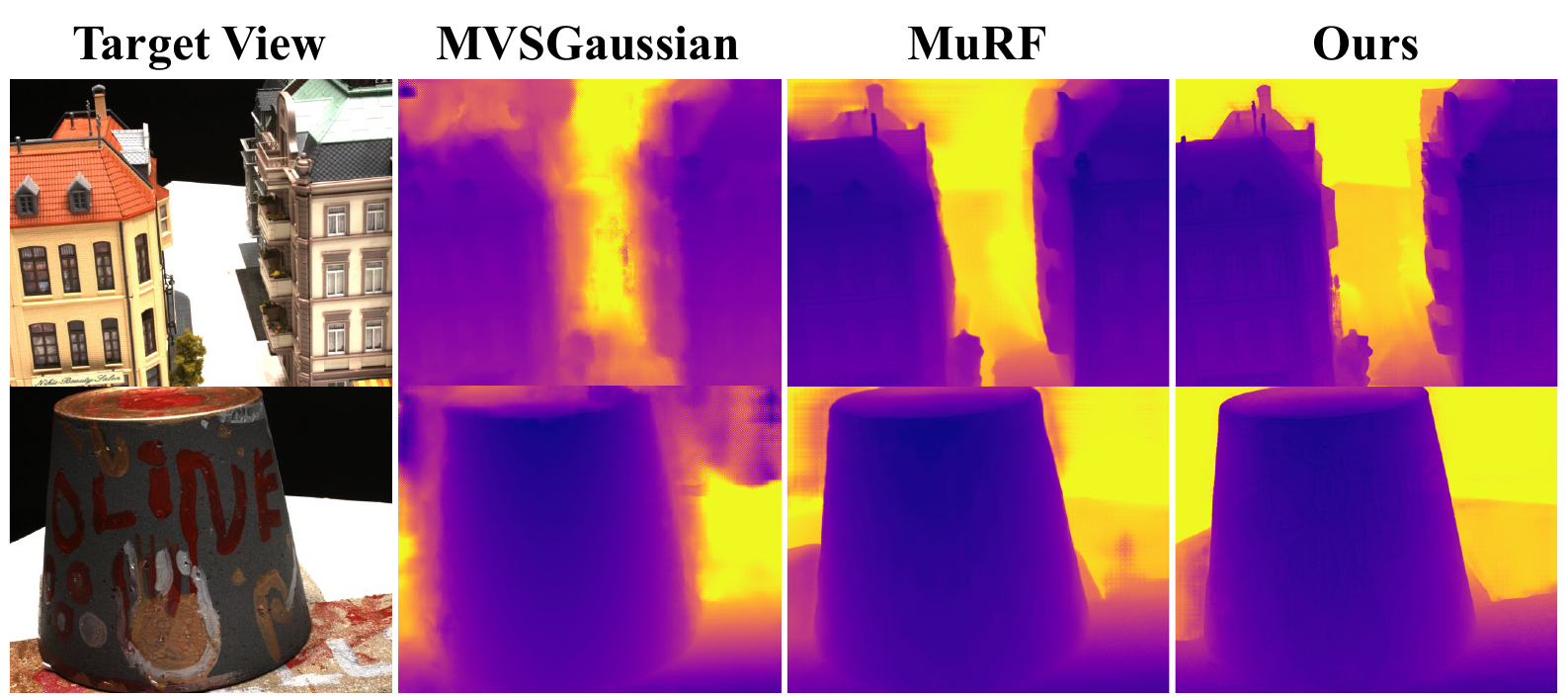}
\vspace{-20pt}
\caption{\textbf{2-view depth prediction on DTU}. Our method yields better detailed geometric information on novel views. } 
\label{fig:depth}
\vspace{-15pt}
\end{figure}

\noindent\textbf{Depth Accuracy on DTU}. To evaluate the quality of reconstructed geometry, we select the DTU dataset as it provides the ground truth of depth. We use quantitative metrics, including the average absolute error ``Abs err'' \cite{yao_mvsnet_2018} and the percentage of pixels with an error less than X mm, denoted as ``Acc(X)'' \cite{liu_mvsgaussian_2024}. As shown in \cref{tab:depth}, our method recovers depth with higher accuracy than the previous best method MVSGaussian \cite{liu_mvsgaussian_2024} in novel view while achieving close accuracy in the reference view. Compared to MuRF \cite{xu_murf_2024}, which prioritizes rendering quality at the expense of geometry accuracy, our method effectively balances both aspects. Moreover, due to the volume rendering approach and implicit representation, MuRF cannot estimate the depth map for reference view since its volume is constructed on the target view, while our 3D-GS-based method can directly generate high-quality reference depth maps. The visual results of the 2-view setting are shown in \cref{fig:depth}. MVSGaussian fails to recover accurate depth due to insufficient context, while our method achieves better details than MuRF.
\begin{figure}[htbp]
\centering
\includegraphics[width=1.0\linewidth]{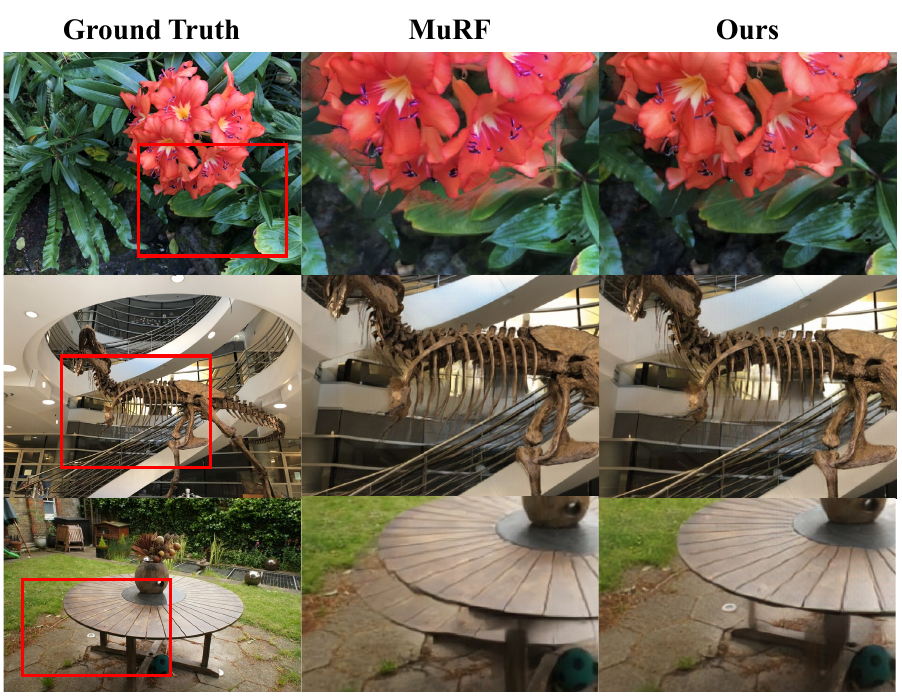}
\vspace{-20pt}
\caption{\textbf{Generalization performance}. The 1st and 2nd rows are from LLFF dataset with 3 input views, and the 3rd row is from the Mip-NeRF 360 Dataset with 2 input views. Fewer artifacts and blurry areas in our results than MuRF.} 
\label{fig:llff_mip}
\vspace{-15pt}
\end{figure}
\begin{figure*}[t]
\centering
\includegraphics[width=1.0\linewidth]{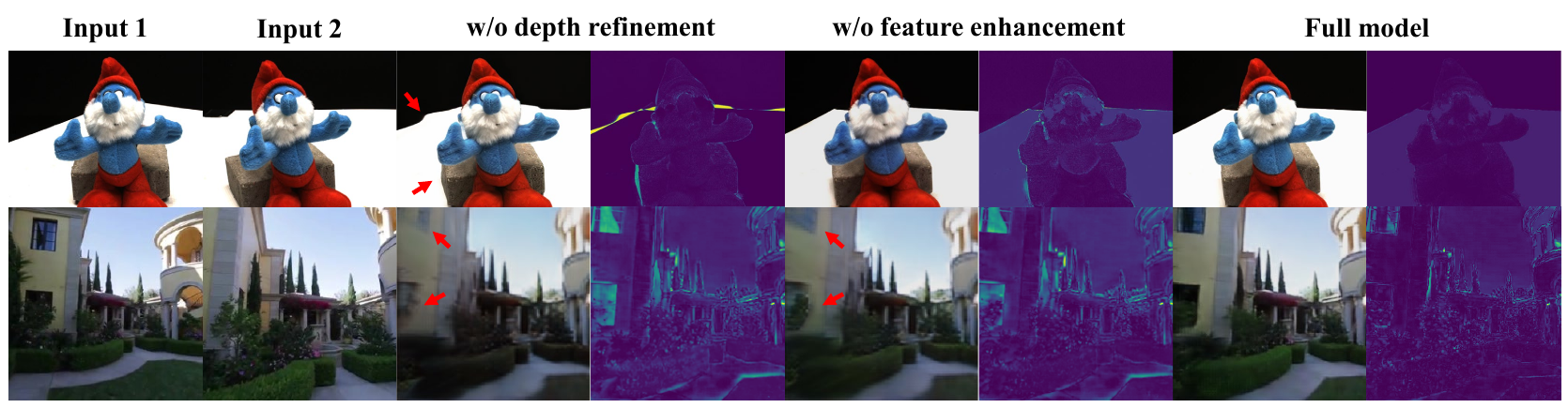}
\vspace{-20pt}
\caption{\textbf{Ablation of fusion strategy}. The 1st row is from small-baseline dataset DTU, and the 2nd row is from large-baseline dataset RealEstate10K. Our proposed fusion strategy works well with the area with occlusion or out of overlap.} 
\label{fig:fusion}
\vspace{-15pt}
\end{figure*}

\noindent\textbf{Generalization Performance}. We also compare the generalization ability of the model trained on large-baseline or small-baseline datasets. In \cref{tab:llff}, we evaluate zero-shot performance on the LLFF dataset \cite{mildenhall_nerf_2020}. All models are trained on the DTU dataset. Our method achieves better scores in PSNR and SSIM, and close scores in LPIPS, compared with the small-baseline method MVSGaussian in a 3-view setting. Regarding the 2-view setting, our method outperforms others in all metrics, showing robustness in handling limited contextual information.

For models trained on RealEstate10K dataset \cite{zhou_stereo_2018}, we evaluate them in both DTU \cite{jensen_large_2014} and Mip-NeRF 360 dataset \cite{barron_mip-nerf_2022}, which is challenging since large-baseline training dataset provides limited supervision for the MVS pipeline due to insufficient overlap between views. The further challenge is that the small-baseline test dataset demands the model to recover precise geometry to obtain high-quality rendering results, which is inherently difficult for large-baseline methods like AttnRend \cite{du2023learning} and MVSplat \cite{chen_mvsplat_2024}. As shown in \cref{tab:mip}, multi-baseline methods, including MuRF \cite{xu_murf_2024} and ours, outperform the specific-baseline methods by a large margin on both datasets, while our method demonstrates even better results compared with MuRF. The visual results shown in \cref{fig:llff_mip} further indicate that our method yields sharper outputs with fewer artifacts than MuRF.

\begin{table}[thbp]
  \centering
  \caption{\textbf{Ablation study on each component of our method.}}
  \vspace{-10pt}
  \resizebox{\linewidth}{!}
    {\begin{tabular}{lccccccc}
    \toprule
    \multirow{2}[4]{*}{Module} & \multicolumn{3}{c}{DTU (Small-Baseline)} &       & \multicolumn{3}{c}{RealEstate10K (Large-Baseline)} \\
\cmidrule{2-4}\cmidrule{6-8}          & PSNR$\uparrow$ & SSIM$\uparrow$ & LPIPS$\downarrow$ &       & PSNR↑ & SSIM↑ & LPIPS↓ \\
    \midrule
    w/o feature enhancement & 27.35 & 0.955 & 0.087 &       & 24.62 & 0.870 & 0.159 \\
    w/o depth refinement & 27.28 & 0.954 & 0.087 &       & 24.56 & 0.869 & 0.161 \\
    w/o reference loss & 27.52 & 0.957 & 0.085 &       & 24.67 & 0.871 & 0.156 \\
    Full model & \textbf{27.56} & \textbf{0.958} & \textbf{0.084} &       & \textbf{24.82} & \textbf{0.873} & \textbf{0.153} \\
    \bottomrule
    \end{tabular}%
    }
  \label{tab:ablation}%
  \vspace{-10pt}
\end{table}%

\subsection{Ablation and Analysis}
As shown in Table~\ref{tab:ablation} and Fig.~\ref{fig:fusion}, we conduct ablation experiments under both large-baseline and small-baseline settings to assess the effectiveness of our designs. The models are evaluated after training separately on DTU \cite{jensen_large_2014} and RealEstate10K \cite{zhou_stereo_2018}.

\noindent\textbf{Depth Refinement}. Our method integrates the source view aligned MDE depth maps into the depth probability volume constructed by the MVS pipeline. To assess the contribution of this depth refinement process, we remove the project-and-sample step along with the subsequent 3D U-net, using the coarse probability volume directly for predicting the target depth map and feature map. As shown in \cref{tab:ablation}, omitting depth refinement leads to a performance decline, which proves that the refined depth prediction indeed improves the final results. The visual comparison in \cref{fig:fusion} further indicates that our depth refinement is helpful to retrieve better geometry, particularly in occluded or low-overlap regions.

\noindent\textbf{Feature Enhancement}. We further explore the difference between the features enhanced by pre-trained monocular features or not. The quantitative results in \cref{tab:ablation} demonstrate that performance drops when only features encoded by the MVS encoder are used for rendering. This suggests that useful inductive bias introduced by monocular features is important for rendering quality. This can be verified in \cref{fig:fusion}, as inaccurate color and texture artifacts emerge when the feature enhancement process is removed.

\begin{figure}[t]
\centering
\includegraphics[width=1.0\linewidth]{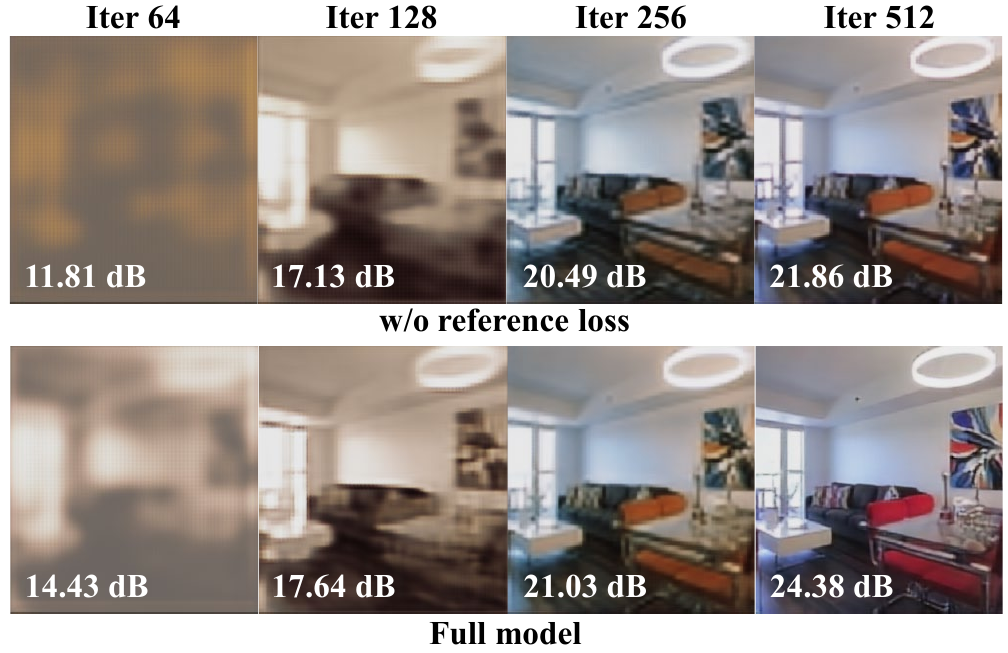}
\vspace{-20pt}
\caption{\textbf{Ablation of reference Loss}. This loss not only elevates the rendering quality, but also boosts the optimization process.} 
\label{fig:loss}
\vspace{-15pt}
\end{figure}

\noindent\textbf{Training Loss}. Our model is trained with an additional reference loss. To evaluate its effectiveness, we separately conduct the training with and without the reference loss under identical settings. As shown in \cref{tab:ablation}, removing the reference loss leads to a decline in final performance, particularly in large-baseline scenarios. Moreover, the visual comparison in \cref{fig:loss} indicates that the reference loss efficiently accelerates the optimization process, yielding more than $2$dB PSNR improvement at specific iterations.

\section{Conclusion}
\label{sec:conclude}

We present \nickName, a feed-forward, generalized 3D Gaussian Spatting approach for novel view synthesis that effectively generalizes across diverse baseline settings. Specifically, we leverage both Multi-View Stereo (MVS) and Monocular Depth Estimation (MDE) to infer depth and enhance the MVS feature with the powerful pre-trained MDE feature. To take advantage of the precision of MVS and the robustness of MDE, we propose a projection-and-sampling mechanism for depth fusion and refine the depth probability volume. To further introduce induction bias for better generalization, we introduce a novel loss function proposed to assist in better geometry and rendering quality.  Experiments shows \nickName\ achieves better multi-baseline generalization as well as better zero-shot performance, proving the effectiveness of our method.

\noindent\textbf{Limitations}. As our method relies on MVS and MDE for depth estimation, it inherits limitations from both, such as decreased depth accuracy in areas with weak textures or specular reflections, resulting in degraded view quality.
\newpage
{
    \small
    \bibliographystyle{ieeenat_fullname}
    \bibliography{main}
}

\end{document}